\documentclass[runningheads]{llncs}
\usepackage{graphicx}
\usepackage{times}
\usepackage{soul}
\usepackage{url}
\usepackage[hidelinks]{hyperref}
\usepackage[utf8]{inputenc}
\usepackage[small]{caption}
\usepackage{graphicx}
\usepackage{amsmath}
\usepackage{booktabs}
\usepackage{algorithm}
\usepackage{algorithmic}
\usepackage[switch]{lineno}
\usepackage{amssymb}
\usepackage{subfig}
\usepackage{siunitx}
\usepackage{booktabs}
\usepackage{placeins}
\usepackage{newfloat}
\usepackage{listings}

\newcommand{\T}{{\bf x}}

\begin{document}

\title{Practical First-Order Bayesian Optimization Algorithms}
\author{Utkarsh Prakash\inst{1}, Aryan Chollera \inst{1}, Kushagra Khatwani \inst{1},  Prabuchandran K. J.\inst{1} \and Tejas Bodas\inst{2}}

%\authorrunning{K. Jain et al.}

\institute{Indian Institute of Technology, Dharwad \and International Institute of Information Technology, Hyderabad \\ \email{tejas.bodas@iiit.ac.in}  } 
%\\ \email{prabukj@iitdh.ac.in}}

\footnotetext[1]{Prabuchandran K.J. was supported by the Science and Engineering Board (SERB), Department of Science and Technology, Government of India for the startup research grant ‘SRG/2021/000048’.}

\maketitle

\begin{abstract}
    First Order Bayesian Optimization (FOBO) is a sample efficient sequential approach to find the global maxima of an expensive-to-evaluate black-box objective function by suitably querying for the function and its gradient evaluations. Such methods assume Gaussian process (GP) models for both, the function and its gradient, and use them to construct an acquisition function that identifies the next query point. In this paper, we propose a class of practical FOBO algorithms that efficiently utilizes the information from the gradient GP to identify potential query points with zero gradients. We construct a multi-level acquisition function where in the first step, we optimize a lower level acquisition function with multiple restarts to identify potential query points with zero gradient value. We then use the upper level acquisition function to rank these query points based on their function values to potentially identify the global maxima. As a final step, the potential point of maxima is chosen as the actual query point. We validate the performance of our proposed algorithms on several test functions and show that our algorithms outperform state-of-the-art FOBO algorithms. We also illustrate the application of our algorithms in finding optimal set of hyper-parameters in machine learning and in learning the optimal policy in reinforcement learning tasks.
\end{abstract}

\section{Introduction}
Zeroth Order Bayesian Optimization (ZOBO) or (BO) is a black-box optimization paradigm to determine the global maxima of an unknown function $f$, i.e., to determine
	\begin{equation*}
		\max_{x \in \mathcal{R} \subset \mathbb{R}^d} f(\textbf{x})
	\end{equation*}
    where $\mathcal{R}$ denotes the domain where we want to search for the global maxima \cite{jones1998efficient}. The objective function $f(\textbf{x})$ is  ``black-box" or is ``latent" in the sense that we do not know the exact functional form of $f({\cdot})$ and can only observe the function values at query locations. Furthermore, these observed values are typically corrupted by noise. 
    This is often the case when the function $f(\cdot)$ is ``expensive to evaluate" i.e. each function evaluation can take a lot of time (for example in days) or is financially costly as in case of drug tests \cite{Pyzer18} and dynamic pricing \cite{Jain23}. Function evaluations can also require a lot of computational power which is the case in automatic hyper-parameter tuning for machine learning \cite{bergstra2013making}, \cite{snoek2012practical}, \cite{golovin2017google}, \cite{falkner2018bohb}, \cite{hoang2018decentralized}, \cite{vien2018bayesian}, and in finding the optimal policy in reinforcement learning tasks \cite{brochu2010tutorial}, \cite{vien2018bayesian}, \cite{muller2021local}.
    In such scenarios, we are limited by the number of function queries that we can perform. This is where BO proves to be useful by providing a sample-efficient sequential strategy of querying the objective function to eventually determine the global optima. 
    
    ZOBO involves two key steps, the first one is that of Bayesian modelling of the unknown function using Gaussian Processes (GP) \cite{williams2006gaussian}. This model provides a probability distribution (a Gaussian distribution) over the set of all possible values that the objective function $f(\textbf{x})$ can take at $\textbf{x}$. Furthermore, using appropriate kernel functions, it also provides a joint probability distribution (multivariate Gaussian) on the function values across different locations.
	In the second step, one constructs a utility or acquisition function from this Gaussian distribution which acts as surrogate for the unknown function. Typical examples of acquisition functions used in the literature are the probability of improvement (PI) \cite{kushner1964new}, expected improvement (EI) \cite{huang2006global}, \cite{jones1998efficient}, \cite{picheny2013quantile}, \cite{mockus1978toward}, upper confidence bound (UCB) \cite{srinivas2010gaussian}, predictive entropy search \cite{hernandez2014predictive} and knowledge gradient (KG) \cite{scott2011correlated}. The point at which the acquisition function is maximized
	correspond to points that either have high objective function values or have high uncertainty in the surrogate model. The objective function is then queried at this maxima point and the Bayesian model is updated based on the new function evaluations. This process is continued until the allowed budget on function evaluations is exhausted. See \cite{williams2006gaussian} and \cite{brochu2010tutorial}, \cite{frazier2018tutorial}, \cite{shahriari2015taking} for more details on GP and BO respectively.
	
	ZOBO has been applied successfully in a variety of domains. In \cite{deshwal2022bayesian}, the problem of optimizing black-box functions  where the input space consists of all permutations of $d$ objects has been considered. In \cite{nguyen2020bayesian}, an attempt to unify the framework of BO with multi-armed bandit problem is proposed to optimize functions in a hybrid input space of continuous and categorical inputs. In \cite{wang2013bayesian}, random embeddings have been utilized to apply BO to high dimensional functions.  \cite{mutny2018efficient},\cite{han2021high} provide alternative approaches to utilize BO for higher dimensional functions by assuming additive structures. More recently, \cite{white2021bananas} has utilized BO for the neural architecture search problem.  
	 
    First Order Bayesian Optimization (FOBO) algorithms \cite{wu2017bayesian}, \cite{prabuchandran2021novel} on the other hand make use of both the function and gradient evaluations at the query points in search for maxima. Note that the gradient information has been utilized extensively in popular optimization procedures such as gradient descent and can also prove to be valuable in BO. Recently, \cite{maclaurin2015gradient}, \cite{franceschi2017forward} and \cite{bohdal2021evograd} have used hyper-gradients for hyper-parameter optimization in machine learning algorithms and \cite{schulman2017proximal}, \cite{sutton2000policy} have constructed the gradient information using the policy gradient method to determine the optimal policy in reinforcement learning. The gradient information is therefore either readily available or can be easily constructed in most applications of our interest for little to no additional cost. In such scenarios,  \cite{wu2017bayesian} and \cite{prabuchandran2021novel} have successfully illustrated that FOBO methods significantly outperform existing ZOBO methods. In fact, \cite{wu2017bayesian} have demonstrated the utility of FOBO methods for hyper-parameter optimization, while \cite{prabuchandran2021novel} have illustrated its use in reinforcement Learning (RL) tasks. In a more recent work, \cite{shekhar2021significance} quantify the improvement in regret using FOBO methods under suitable assumptions and propose a FOBO algorithm that achieves a superior regret of $O((\log n)^{2})$ compared to $\Omega(\sqrt{n})$ achieved by ZOBO. 
    
Current FOBO methods, even though efficient in their search, have major shortcomings. A major computational challenge in \cite{wu2017bayesian} in utilizing the gradient information is in performing the posterior updates to the GP. This involves taking an inverse of an $n(d+1)$ dimensional kernel matrix, requiring  $O((n(d+1))^3)$ computations. In order to alleviate this problem \cite{prabuchandran2021novel} model each partial derivative using an independent Gaussian process which reduces the computational complexity to $O((d+1)n^{3})$. They use an acquisition function, which independently search in each dimension for points in the domain with zero partial derivative. While promising, this approach however need not necessarily recommend query points with zero gradient and hence the gradient information is not efficiently utilized. The proposed algorithm also does not utilize the uncertainties in the estimates of the objective function (which is easily available and valuable in such Bayesian modelling) and therefore fails to explore the search space efficiently. 

In this paper, we propose FOBO algorithms where the impetus is to search for potential query points in the domain where the gradient vanishes. We assume independent GP models for each of the partial derivatives and construct a novel multi-level acquisition function. Using the notion of multiple restarts, we first optimize the lower level acquisition function to identify potential query points where the partial derivatives are zero across all dimensions simultaneously. Such points with zero gradient could potentially correspond to maxima, minima or saddle point and hence we need to rank them on the basis of their likelihood of being the global maxima. Towards this, we use an upper level acquisition (significance) function that makes use of the GP model of the objective function to rank these points. The point with the highest significance value is chosen as the next query point. 
%It is important to note that, none of the prior FOBO algorithms are capable of making this distinction.
In addition to this, to encourage exploration in search for maxima, we utilize uncertainties in the estimates of the function and the partial derivative values. Our algorithms are broadly categorized into two types, the first one is based on the EI algorithm while the second one is based on PI. We illustrate their performance on standard test functions and show its efficacy against the state of the art ZOBO and FOBO algorithms. We also consider their application  in automatic hyper-parameter tuning for machine learning and finding optimal policy in reinforcement learning.

\section{First Order Bayesian Optimization}
\label{fobo}
Recall that FOBO methods utilize the partial gradient information along with the function values at the query points to search for global optima. As discussed earlier, let $f$ be the function to be optimized in the domain $\mathcal{R}$ where $\mathcal{R} \subset \mathbb{R}^{d}$. FOBO methods place a GP prior over $f : \mathcal{R} \rightarrow \mathbb{R}$ with mean function $\mu : \mathcal{R} \rightarrow \mathbb{R}$ and kernel function $K : \mathcal{R} \times \mathcal{R} \rightarrow \mathbb{R}$ similar to that in ZOBO. \cite{wu2017bayesian} leverage the fact that gradient of a GP is also a GP \cite{williams2006gaussian} and model the tuple $(f(\cdot), \nabla f(\cdot)) \in \mathbb{R}^{d+1}$ as a multi-output Gaussian process with mean function $\tilde{\mu}$ and covariance function $\tilde{K}$ defined as follows:
    \begin{align*}
         \tilde{\mu} &= (\mu(\textbf{x}), \nabla \mu  (\textbf{x}))^{T} \in \mathbb{R}^{d+1}, \nonumber \\
         \tilde{K}(\T,\T')&=
\begin{pmatrix}
K(\T,\T') & J(\T,\T') \\
J(\T,\T')^T & H(\T,\T') 
\end{pmatrix}, ~~\tilde{K} \in \mathbb{R}^{d+1 \times d+1} \nonumber
    \end{align*}
    where $J(\T,\T')=\big{(}\frac{\partial K(\T,\T')}{\partial \T'(1)},\frac{\partial K(\T,\T')}{\partial \T'(2)},\cdots,\frac{\partial K(\T,\T')}{\partial \T'(d)}\big{)}$
and $H(\T,\T')$ is the $d \times d$ dimensional Hessian of $K(\T,\T')$.
As a next step, an appropriate acquisition function is constructed that makes use of this multi-output GP. Due to the additional gradient information in FOBO, the query points suggested by this acquisition function has a higher likelihood of finding the maxima faster compared to ZOBO methods. As outlined earlier, the main disadvantage however, of modelling the function and its gradient as a joint GP is that the posterior updates require taking inverse of an $n(d+1)$ dimensional kernel matrix which requires $O((n(d+1))^3)$ computations where $n$ is the number of function evaluations. On the other hand, ZOBO methods only require taking inverse of an $n$ dimensional matrix which is equivalent to performing $O(n^3)$ computations. This high computational effort required for FOBO severely limits its applicability.
    
 \cite{prabuchandran2021novel} overcome this computational limitation by modelling each partial derivative as an independent GP.  This results in $d+1$ independent GPs (one for the function value and $d$ for each partial derivative), i.e.,
    \begin{equation}
        f(\cdot) \stackrel{}{\sim} \mathcal{GP}(\mu(\cdot), K(\cdot , \cdot))
        \label{fGP}
    \nonumber
    \end{equation}
    
    \begin{equation}
        \frac{\partial f(\cdot)}{\partial \T(i)} \stackrel{}{\sim} \mathcal{GP}(\mu_{i}(\cdot), K_{i}(\cdot , \cdot)), i \in \{1,...,d\}.
        \label{pdGP}
    \nonumber
    \end{equation}
    With this, the computational burden reduces from $O((n(d+1))^3)$ to $O((d+1)n^3)$. Furthermore, as the $d+1$ GPs are independent, their posterior update step can also be parallelized. They propose the following acquisition function for each of the partial derivative GPs:
    \begin{equation*}
        I_{i}(\textbf {x}) = \mathbb{E}_{n} \left( \left |\frac{\partial f(\cdot)}{\partial \textbf{x}(i)} \right | \right ),     i \in \{1,...,d\}
        \label{prabu_acquisition_function}
    \end{equation*}
     where $\mathbb{E}_{n}$ represents the conditional expectation from the posterior GP model given $n$ function values.  The next query point suggested by the $i$th partial derivative GP is then given by,
    \begin{equation*}
        \textbf{x}^{n+1}_{i} =\arg \min_{\textbf{x} \in D} I_{i}(\textbf {x}), i \in {1,...,d}.
        \label{argmin_of_custom_acquisition_function}
    \end{equation*}
    As for the point suggested from the GP model of the objective function, denoted by $\textbf{x}_{0}^{n+1}$, any of the standard acquisition functions such as EI or KG could be used. This procedure leaves us with $d+1$ candidates for the next query point. Note that we still have to decide which of the candidate point should actually be queried. 
    Towards this, \cite{prabuchandran2021novel} suggest the following two alternatives:
    \begin{enumerate}
        \item The information is aggregated by taking a weighted convex combination of all the candidates
            i.e.,
            \begin{equation*}
                \textbf {x}^{n+1} = \sum_{i=0}^{d} \frac {\exp(\mu^{(n)}(\textbf{x}^{n+1}_{i}))}{\sum_{i=0}^{d} \exp(\mu^{(n)}(\textbf{x}^{n+1}_{i}))} \textbf{x}^{n+1}_{i}.
                \label{convex_combination}
            \end{equation*}
        \item The information is aggregated by taking the point that has the highest 
            estimated mean i.e. $\textbf{x}^{n+1} = \textbf{x}^{n+1}_{i^{*}}$ where
            \begin{equation*}
            \begin{gathered}
                i^{*} = \arg \max_{i} \mu^{(n)}(\textbf{x}^{n+1}_{i}), i \in \{0,...,d\}.                .
            \end{gathered}
            \label{maximum_combination}
            \end{equation*}
        \end{enumerate}
    Note that since the acquisition functions are independently optimized, the candidate points only have their partial derivative values in some dimension close to zero. This in no way implies that the gradient value at this point is also zero. Furthermore, the acquisition function does not take into account the uncertainties in the estimates for the partial derivatives. Therefore, points having modulus of the partial derivative close to $0$ with high uncertainty can be suggested.

\section{Proposed Algorithm}\label{proposedFobo}
In this section, we propose two gradient based 
multi-level acquisition functions, namely, gradient based expected improvement (\textit{gEI}) and the gradient based probability of improvement (\textit{gPI}) that use features from existing acquisition functions such that GP-UCB, EI and PI efficiently with the hope of identifying query points with potentially zero gradient. The multi-level acquisition functions have the same functional form at the upper level while differing  only in the lower level. When the context is clear, we will also refer to the lower level acquisition functions by \textit{gEI} and \textit{gPI} respectively. The lower level acquisition functions are based on the EI \cite{huang2006global}, \cite{jones1998efficient}, \cite{picheny2013quantile}, \cite{mockus1978toward}, PI \cite{kushner1964new} and GP-UCB \cite{srinivas2010gaussian} but additionally incorporate the gradient information effectively when suggesting the next query point. Furthermore, each lower level acquisition function is optimized with $k$ different seed points (restarts) and hence, each run of the lower-level acquisition optimizer results in potentially multiple candidates (points with zero gradients) to query. This is where we design and use an upper level acquisition function that ranks the potential query points (using the function GP) according to their likelihood of being points with zero gradient.

\subsection{Gaussian Process Regression}\label{gpr}
Let $\mathcal{D}^{(1:n)} = \{\textbf{x}^{(1:n)}, (\hat{f}^{(1:n)}, \widehat{\nabla f}^{(1:n)})\}$, be the function and gradient evaluations from $n$ iterations. Additionally, let us define $\mathcal{F}^{(1:n)} :=  \{\textbf{x}^{(1:n)}, \hat{f}^{(1:n)}\}$. We have a Bayesian model on the function and each of its partial derivatives and assume that the GP priors are independent. In what follows, we will first discuss the GP regression for the objective function GP. Since the GP models are independent, the regression for the partial derivative GPs follows along similar lines.  To account for noise in the function evaluations, we assume the  $\hat{f}(\T)$ is normally distributed, i.e.,
\begin{align*}\label{NM}
\hat{f}(\T) \big{|} {f}(\T) \sim \mathcal{N}\Big{(}{f}(\T), \lambda^2\Big{)} \nonumber
\end{align*}
where $\lambda^2: \mathcal{R} \rightarrow \mathbb{R}$ specifies the noise variance. We also assume that the noise variance is constant across the domain $\mathcal{R}$.
For specifying the prior, we use the constant mean function $\mu_{0} \in \mathbb{R}$ and the squared exponential function $k$ 
\begin{equation*}\label{squared_exponential_kernel}
			k(\T , \T') = \sigma^2 \exp \left (-\frac{(\T - \T')^T (\T - \T')}{2l^2}\right ).
\end{equation*}
Let the kernel matrix $\textbf{K}$ be defined as
\begin{equation*}
	\textbf{K} := \begin{bmatrix}
				k(\textbf{x}^{1}, \textbf{x}^{1}) & \hdots & k(\textbf{x}^{1}, \textbf{x}^{n})\\
				\vdots & \ddots & \vdots \\
				k(\textbf{x}^{n}, \textbf{x}^{1}) & \hdots & k(\textbf{x}^{n}, \textbf{x}^{n})
				\end{bmatrix}
\end{equation*}
and additionally define $\textbf{K}_{y} := \textbf{K} + \lambda^{2}I$.
The posterior for the objective function $\hat{f}$ at any $\T$ \cite{williams2006gaussian} is then given by
\begin{equation}\label{postGP}
	P(\hat{f}| \mathcal{F}^{1:n}, \textbf{x}) = \mathcal{N} (\mu^{(n)}(\textbf{x}), \sigma^{(n)}(\textbf{x}) + \lambda^{2})
\end{equation}
where,
\begin{eqnarray*}
	\mu^{(n)}(\textbf{x}) &=& \mu_0 + \textbf{k}^{T}\textbf{K}_{y}^{-1}(\hat{f}^{1:n}-\mu(\T^{1:n})) \\
	\sigma^{(n)}(\textbf{x}) &=& k(\textbf{x}, \textbf{x}) - \textbf{k}^{T}\textbf{K}_{y}^{-1}\textbf{k} \\
	\textbf{k} &=& \begin{bmatrix}
						k(\textbf{x}, \textbf{x}^{1}) & k(\textbf{x}, \textbf{x}^{2}) & \hdots & k(\textbf{x}, \textbf{x}^{n})				
						\end{bmatrix}.
%\end{gathered}
\end{eqnarray*}
Similar expressions can be obtained for the partial derivative GPs as well. Let  $\mu^{(n)}_{i}(\T)$ and  $\sigma^{(n)}_{i}(\T) + \lambda^{2}_{i}$ denote the mean and standard deviation of $\frac{\partial f(\cdot)}{\partial x(i)}$ after $n$ posterior updates. 
The variables $\eta=\{\sigma^2,l,~\lambda^2,\mu_0\}$ are referred to as the hyper-parameters of the GP model \cite{williams2006gaussian}.
Note that for each choice of these hyper-parameters, we obtain a different prior distribution over the function and one must tune these hyper-parameters so that they are consistent with the observed data. The hyper-parameters can be estimated using either maximum likelihood estimation (MLE), maximum a posteriori (MAP) or a fully Bayesian approach \cite{frazier2018tutorial}, \cite{snoek2012practical}. In our work, we utilize the MLE approach  instead of the Bayesian methods as they require MCMC sampling methods \cite{snoek2012practical} which are computationally very expensive. MLE involves determining the optimal set of hyper-parameters that maximizes the likelihood of the observed data, i.e.,
\begin{equation*}
    \eta^{*} = \arg \max P(\hat{f}^{1:n}|\eta) \mbox{~where~}
\nonumber
\end{equation*}
\begin{equation}
\nonumber
    \log P(\hat{f}^{1:n}|\eta) = -\frac{1}{2}\hat{f}^{{(1:n)}^T}\textbf{K}_{y}^{-1}\hat{f}^{(1:n)} -\frac{1}{2}\log|\textbf{K}_{y}| + C
\end{equation}
where $C=-\frac{n}{2}\log 2\pi$.
This process of determining the best hyper-parameters by maximizing the log likelihood of the data is referred to as ``GP fitting". As a first step, we perform this GP fitting for the function as well the partial derivative GPs using the observed data $\mathcal{D}^{(1:n)}$. Using the optimal hyper-parameters, we then obtain the posterior GP models using \eqref{postGP}.
\subsection{Multi-level \textit{gEI} Acquisition Function}\label{gEI}
In this subsection, we discuss the construction of our \textit{gEI} acquisition function from the posterior GP models. A natural way to overcome the limitations of the FOBO algorithm  \cite{prabuchandran2021novel} is to combine information from all partial derivatives effectively and search for points with zero gradient directly. Towards this, we define the utility function for each partial derivative model as follows. 
\begin{equation}
    I_{i}(\textbf {x}) = \mathbb{E}_{n} \left( \left |\frac{\partial f(\cdot)}{\partial x(i)} \right | \right ) +  \sigma_{n} \left( \left |\frac{\partial f(\cdot)}{\partial x(i)} \right | \right ),   i \in \{1,...,d\}
    \label{custom_gradient_acquisition_function_plus_variance}
\end{equation}
where $\mathbb{E}_{n}$ and $\sigma_{n}$ represents the conditional expectation and standard deviation calculated using $i^{th}$ partial derivative GP model after $n$ posterior updates respectively. Since, $\frac{\partial f(\cdot)}{\partial x(i)}$ is a Gaussian random variable, the utility function expression can be calculated using the following Lemma.
\begin{lemma}
\label{expectation_of_folded_normal}
Let Z be a Gaussian random variable with mean $\mu$ and standard deviation $\sigma$. Then we have $\mathbb{E}(|Z|) = 2 \sigma \phi(-\frac{\mu}{\sigma}) + \mu\left[ \Phi(\frac{\mu}{\sigma}) - \Phi(-\frac{\mu}{\sigma}) \right ]$ and $\sigma(|Z|) = \sqrt{\mu^{2} + \sigma^{2} - \mathbb{E}(|Z|)}$ 
where $\phi(\cdot)$ and $\Phi(\cdot)$ correspond to the standard normal probability density function and cumulative density function respectively. (Proof in the supplementary material).
\end{lemma}
In order to determine points where all partial derivatives are zero, we define our lower acquisition function as a sum of the utility functions of all the partial derivatives. Intuitively, this also gives us a good measure of the gradient value at that point. %how close each partial derivative is to $0$ in all the dimensions.
Our lower level acquisition function (also named as \textit{gEI}) is given by
\begin{equation}
    \textit{gEI}(\T) = \sum_{i=1}^{d} I_{i}(\textbf {x})
    \label{combineDAcq}
\end{equation}
and the next potential candidate point is given by
\begin{equation*}
    \textbf{x}^{n+1}_{D} = \arg \min_{\textbf{x} \in \mathcal{R}}{\textit{gEI}(\T)}.
\end{equation*}
Note that $gEI(\T)$ could be a multi-modal function and therefore,  can have multiple global minima, local minima, and/or saddle points. Of these, it is only the global minima corresponds to optima of the objective function (either maxima, minima or saddle point). It is therefore of pertinent interest to identify all such global minimas of $gEI(\T)$ and therefore as many optima (maxima, minima or saddle points) of the objective function as possible. %Therefore, on optimizing $gEI(\T)$ once, we are not guaranteed to always find the maxima with respect to the objective function. Secondly, while optimizing $gEI(\T)$, standard optimization algorithms may converge to local minima instead of a global one.
%To alleviate these problem,
To accomplish this goal, we propose to restart the optimization of $gEI(\T)$ with $k$ different seed points. By doing so, we construct a rich set of $k$ candidate points, where at most $k$ 
of them could now be the global minima of $gEI(\T)$. Let us denote these candidate next points by $\textbf{x}^{n+1}_{i}, ~~i \in \{1, ..., k\}$, corresponding to the point obtained from each of the restarts. We now have at most $k$ potential optima points of the objective function and what remains is to segregate them into maxima, minima or saddle point. We utilize the significance of a point defined as follows
\begin{equation}
    s(\T) = \mu^{(n)}(\T) + \alpha\sigma^{(n)}(\T)
    \label{significance}
\end{equation}
to distinguish between potential points that could correspond to maxima, minima and saddle point w.r.t the objective function. High value for $\mu^{(n)}(\cdot)$ ensures exploitation whereas high value for $\sigma^{(n)}$ aids in exploration. Moreover, we would initially want our algorithm to explore the search space to gather information about different points and would later want the algorithm to exploit from the information thus obtained. To achieve this, one can vary $\alpha$ as a function of $n$ to strike-off a balance between exploration and exploitation.
Note that the $d$ points suggested in the FOBO algorithm by \cite{prabuchandran2021novel} and $k$ points suggested by our algorithm are qualitatively very different. The points suggested in our algorithm correspond to having gradients close to zero whereas in the algorithm proposed by \cite{prabuchandran2021novel}, the points may not have partial derivatives close to zero all at once along all dimensions.
We also use  EI  for identifying the potential query point as suggested by the objective function GP. Let this point be represented by $\textbf{x}_{fGP}^{n+1}$.
Now define the set of potential points as $Q := \{\textbf{x}_{1}^{n+1},..., \textbf{x}_{k}^{n+1},  {\T}_{fGP}^{n+1}\}$. Using the definition for significance of a point, we use one of the following methods to identify the next query point and this forms the upper-level of the proposed acquisition function.
\begin{enumerate}
    \item \textbf{Maximum Significance (MS) }: The next query point suggested by our algorithm is
    \begin{equation}
        \textbf {x}^{n+1} = \arg \max_{\T \in Q} s(\T)
        \label{maximum_significance}
    \end{equation}
    \item \textbf{Maximum Significance with Convex point (MSC) }:  We take convex combination of points in Q as follows:
    \begin{equation*}
        \textbf {x}^{n+1}_{convex} = \sum_{\T \in Q} \frac {\exp(s(\T))}{ \sum_{\T \in Q} \exp(s(\T))} \textbf{x}
    \end{equation*}
    Let $ Q' = Q \cup {\textbf {x}^{n+1}_{convex}}$. Then the next query point suggested by our algorithm is
    \begin{equation}
        \textbf {x}^{n+1} = \arg \max_{\T \in Q'} s(\T)
        \label{maximum_significance_convex}
    \end{equation}
\end{enumerate}
We will use \textit{gEI} to refer to acquisition function as well as the algorithm interchangeably wherever the context is clear.

\subsection{Multi-level \textit{gPI} Acquisition Function}
The \textit{gEI} acquisition function discussed in the previous section identifies the next query point in two steps. First, we obtain potential points that have zero gradient and then filter it further (depending on whether they are maxima, minima or saddle point) to obtain the next query point. This two step process can be simplified if all the information can be combined at once.
In what follows, we propose \textit{gPI}, an acquisition function based on probability of improvement (PI) \cite{kushner1964new} where we incorporate the function values in the acquisition function. \textit{gPI} can now directly search for the maxima rather than relying on \eqref{significance} to distinguish between minima, maxima and saddle points.
In \textit{gPI}, since all GPs are independent, we search for points where the probability of having all the partial derivatives close to zero is high. We define the utility of  $\frac{\partial f(\cdot)}{\partial x(i)}$ as the probability that the partial derivative at $\T$ lies in the range $[-\epsilon, \epsilon]$ i.e.,
\begin{equation*}
    P_i(\T) = P\left(  \frac{\partial f(\cdot)}{\partial x(i)} \in (-\epsilon,\epsilon) \right ), i \in \{1,...,d\}.
\nonumber
\end{equation*}
Since, $\frac{\partial f(\cdot)}{\partial x(i)}$ is Gaussian, the above expression reduces to
\begin{equation*}
    P_i(\T) = \Phi \left ( \frac{\epsilon - \mu^{(n)}_{i}(\T)}{\sigma^{(n)}_{i}(\T) + \lambda^{2}_{i}}\right ) - \Phi \left ( \frac{-(\epsilon + \mu^{(n)}_{i}(\T))}{\sigma^{(n)}_{i}(\T) + \lambda^{2}_{i}}\right ).
\nonumber
\end{equation*}
In order to search for such points where the partial derivative is close to zero in all dimensions, we define the utility of the gradient as the product of $P_i(\T)$ over the $d$ dimensions.
In order to incorporate the function value in the acquisition function, we utilize the standard PI acquisition function \cite{kushner1964new} with an exploration factor $\epsilon$.
Since, $f(\T)$ is Gaussian with mean $\mu^{(n)}$ and standard deviation $\sigma^{(n)}$ after $n$ posterior updates, the PI acquisition is given by $P_0(\T) = P(f(\T) \geq f(\T^{+}) + \epsilon)$, where $\T^{+}$ denotes the best point seen so far. From this it is easy to see that,
\begin{equation*}
    P_0(\T) = \Phi \left ( \frac{\mu^{(n)}(\T) - f(\T^{+}) - \epsilon}{\sigma^{(n)}(\T) + \lambda^{2}} \right )
\end{equation*}
The lower-level \textit{gPI} acquisition function is now defined as  
\begin{equation}
    \textit{gPI}(\T) := \prod_{i=0}^{d} P_i(\T)
    \label{gPI_combined}
\end{equation}
and the next query point is given by
\begin{equation*}
    \textbf{x}^{n+1}_{D}= \arg \max_{\textbf{x} \in \mathcal{R}}{\textit{gPI}(\T)}
\end{equation*}
% \textcolor{red}{Yu can call the function PI term as $P_0(\T)$ and merge it into product term.}
Similar to \textit{gEI} we optimize the above acquisition function with $k$ different restarts to obtain several candidate points that are potential points of maxima. We then
 combine them similar to \textit{gEI} acquisition either using Maximum Significance (MS) or Maximum Significance with Convex Point (MSC) which forms the upper-level procedure.
 Algorithm \ref{pseudo_code} gives a complete description our proposed FOBO algorithms that utilize \textit{gEI} or \textit{gPI} acquisition functions.
 \begin{algorithm}[ht]
\caption{\textit{gEI} / \textit{gPI} algorithm}\label{pseudo_code}
\begin{algorithmic}
\STATE \textbf{Input:}

%$ \text{Initialdata}\leftarrow D_{1:n_0} = \{\T^{i}, \hat{\rho}(\T^{i}), \widehat{\nabla\rho}(\T^{i}), \tau_{i}\}|_{i=1}^{n_0}$

Budget $B$, initial function and gradient evaluations $(\hat{f}(\cdot),\widehat{\nabla f(\cdot)})$

%\\ type $\leftarrow$ Type of the acquisition function

%\STATE \textbf{Output:} $\T^{*} \approx \arg\max_{\T} f(\T)$

%\STATE Set up prior using the data $D_{1:n_0}$ obtained at $n_0$ arbitrary initial points
\STATE \textbf{Algorithm:}

%\STATE $\mathcal{D}_0=\{ \emptyset \}$

\FOR {$n=0, 1, 2...B-1$} 

\STATE Find the new point $\T^{n+1}_{fGP}$ suggested by the function GP using EI acquisition function.

\STATE Find $k$ points by minimizing \eqref{combineDAcq} for \textit{gEI} (\eqref{gPI_combined} for \textit{gPI}).

\STATE Find the next query point $\T^{n+1}$ using the MS or MSC method (see \eqref{maximum_significance} and \eqref{maximum_significance_convex}).

\STATE Get $(\hat{f}(\T^{n+1}),\widehat{\nabla f}(\T^{n+1}))$ at $\T^{n+1}$

\STATE Augment the data $\{(\hat{f}(\T^{n+1}),\widehat{\nabla f}(\T^{n+1}))\}$ into the set of function and gradient evaluations.

\STATE Update all the GPs using the augmented data.

\ENDFOR

\RETURN $\T^*=\arg\max \hat{f}^{1:n}$

\end{algorithmic}

\end{algorithm}

\section{Performance Evaluation and Results}
In this section, we compare the performance of \textit{gPI}, \textit{gEI}, ZOBO (specifically, EI) and the state-of-art FOBO algorithm of \cite{prabuchandran2021novel} on 6 synthetic test functions, 2 RL tasks and 1 hyper-parameter tuning task. We refer \textit{gPI} algorithm with MS and \textit{gPI} with MSC as \textit{gPI-MS} and \textit{gPI-MSC} respectively. Similar notations are used for \textit{gEI}. Henceforth, we refer to the algorithm proposed by \cite{prabuchandran2021novel} as FOBO. We leverage the BoTorch framework \cite{balandat2020botorch} for performing our experiments and use their built-in optimizers for GP fitting and optimization of acquisition function. We average the results over multiple runs of the algorithms which we parallelize using Multi Processing Interface (MPI).

\begin{figure*}
\begin{center}
 \subfloat[Branin]{
  	\includegraphics[width = 5.5cm]{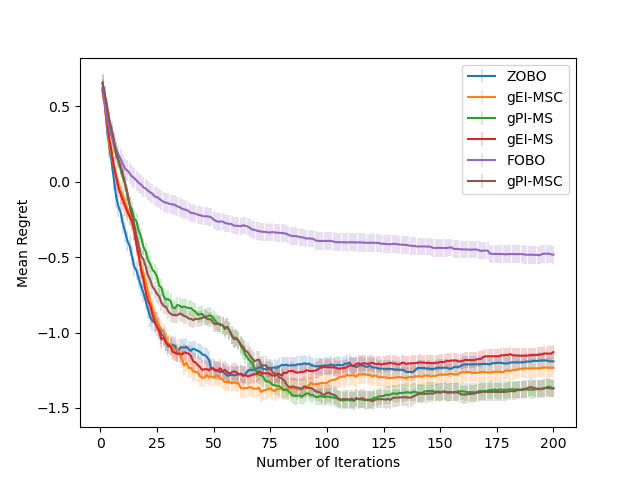}
  	
  	}
    \hspace{1mm}
    \subfloat[Levy]{
  	\includegraphics[width=5.5cm]{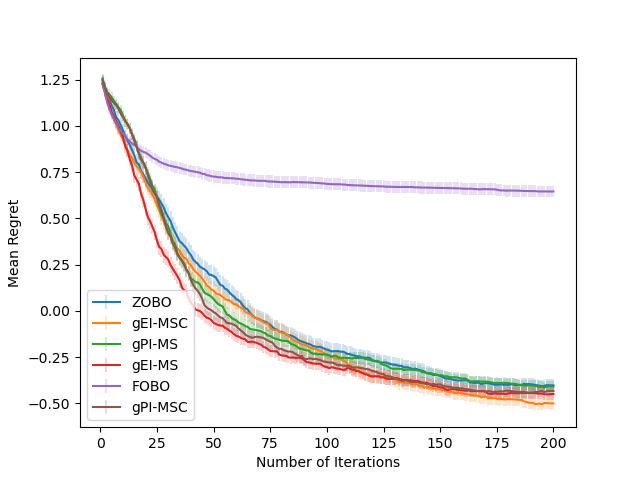}
  	
  	}
   \hspace{0.25mm}
 \subfloat[Ackley]{
    \includegraphics[width=5.5cm]{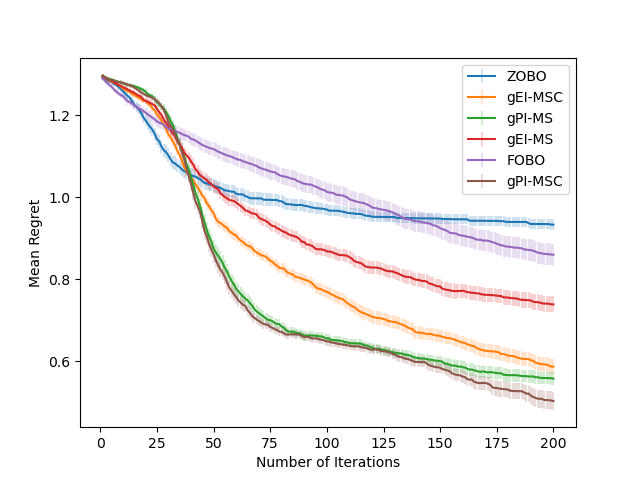}
    
    }
 \\
     \subfloat[DixonPrice]{
    \includegraphics[width=5.5cm]{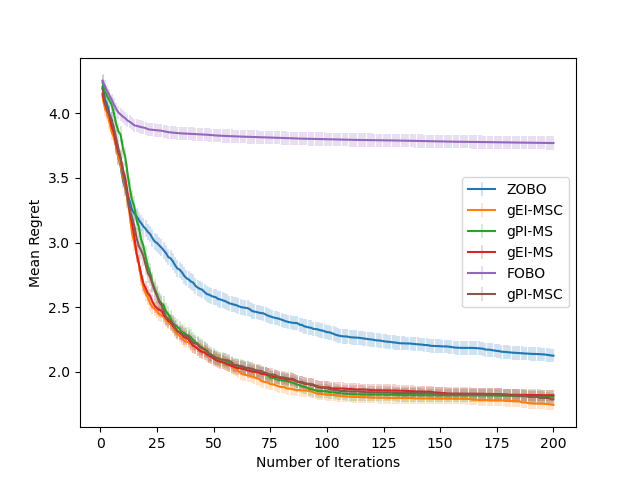}}
    \hspace{1mm}
 \subfloat[Hartmann]{
  	\includegraphics[width=5.5cm]{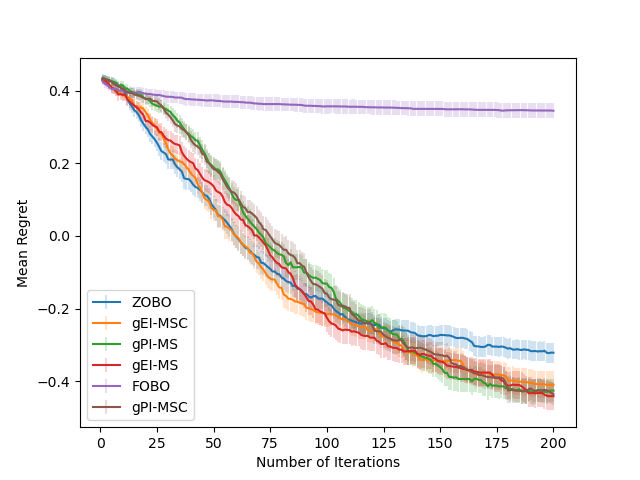}
  	}
\hspace{0.25mm}
\subfloat[Cosine8]{
    \includegraphics[width=5.5cm]{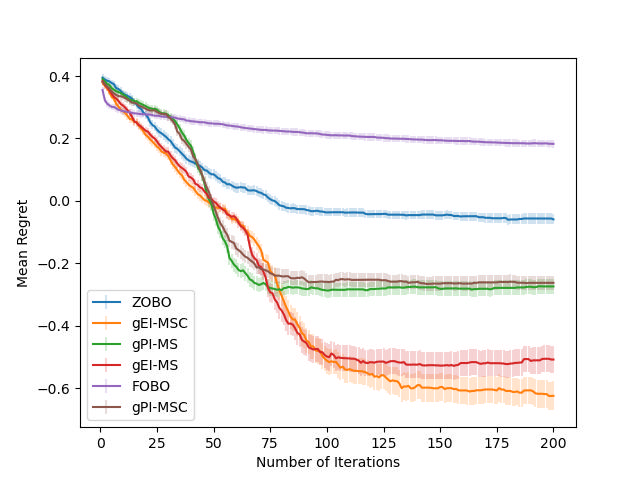}
    }
       \caption{Synthetic Test Functions}
 \label{TestFunctions}
 \end{center}
 \end{figure*}

\subsection{Synthetic test functions}
The 6 synthetic test functions considered for performance evaluation are Branin (dimension of the domain d=2), Levy (d=4), Ackley (d=5), DixonPrice (d=5), Hartmann (d=6) and Cosine8 (d=8) from \cite{simulationlib}. We minimize these functions which is equivalent to  maximizing the negative of the functions. A Gaussian noise with 0 mean and a variance of 0.25 was added to the function and gradient evaluations to test the robustness of the algorithms. We plot the mean of the logarithm of the immediate regret as a function of number of iterations (function evaluations) to compare the performance of the algorithms. Note that the immediate regret is defined as the absolute value of the difference of the maximum of the solutions obtained by the BO method so far and the global maximum. We start with $5$ initial points to fit the GPs and run the algorithms for $200$ iterations. We average the logarithm of immediate regret over $200$ runs to obtain the mean regret. For the \textit{gPI} and \textit{gEI} algorithms, the value of $k$ chosen is 10 and this choice of $k$ is arbitrary. However, $k$ can be treated as a hyper-parameter itself and optimized which is left for future work.
\par
From Figure \ref{TestFunctions}, it is evident that \textit{gPI} and \textit{gEI} algorithms achieve significantly lower regrets for Ackley, DixonPrice, Hartmann and Cosine8 than ZOBO and FOBO. This shows that our algorithms are able to utilize gradient information in a much more effective way and at the same time distinguish between maxima, minima and saddle points. For Branin, the performance of ZOBO is better than \textit{gEI-MS} algorithm. In contrast, \textit{gPI-MS} and \textit{gPI-MSC} achieve lower regret than ZOBO. However, notice that the logarithm of immediate regret is around $-1.5$ for all the algorithms i.e., the error between the true optimal value and that found by our algorithm is less than $0.1$. Therefore, all the algorithms have almost converged to the true optimal value. For Levy, the performance of all the algorithms is similar. For most of the functions, \textit{gPI} class of algorithms perform better or similar to \textit{gEI}. This is because \textit{gPI} takes into account the function value into the acquisition function itself and therefore, is able to suggest better points. However, \textit{gEI} performs much better than \textit{gPI} for Cosine8. Further research is warranted in order to understand where \textit{gEI} is more suitable than \textit{gPI} and vice-versa. %Note that the performance of the FOBO algorithm is poorer than that of ZOBO which is contrary to the results  in \cite{prabuchandran2021novel}. This is primarily because we fit the hyper-parameters using MLE, whereas \cite{prabuchandran2021novel} uses  Bayesian estimation.

\subsection{Hyper-parameter Optimization}
We perform two experiments on hyperparameter optimization and show that our algorithms work well for optimizing hyper-parameters for machine learning tasks. We obtain the hyper-gradients (gradients w.r.t. hyper-parameters) using the EvoGrad approach described in \cite{bohdal2021evograd}. %(Refer supplementary material for results on hyper-parameter tuning with 6D Regularization.)
\begin{figure*}
 \begin{center}
 \subfloat[6D Regularization Problem]{
  	\includegraphics[width = 5.5cm]{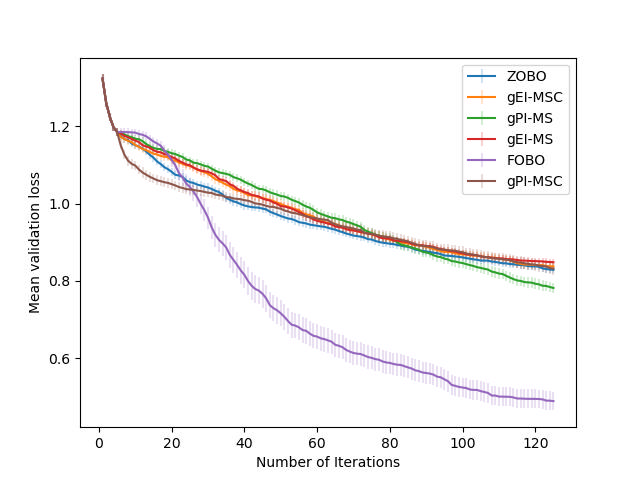}
  	\label{6dreg}
  	}
   \subfloat[Rotation transformation]{
  	\includegraphics[width = 5.5cm]{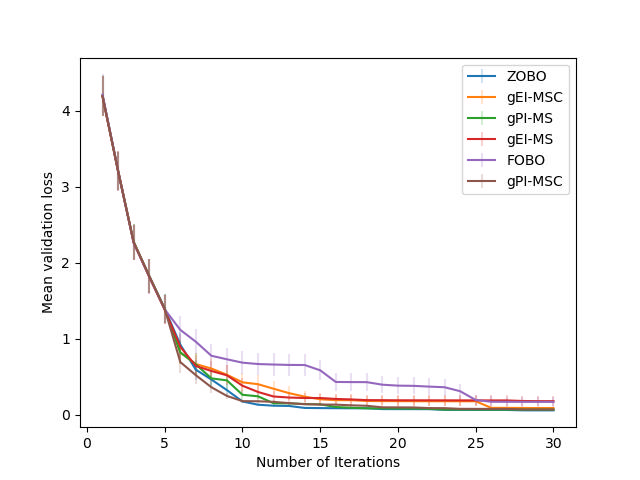}
  	\label{rotTrans}
  	}
  	% \subfloat[Gridworld]{
  	% \includegraphics[width = 5.5cm]{Regret_plot_gw.jpeg}
  	% \label{grid_world}
  	% }
  	\caption{Hyper-parameter Tasks}
  	\label{Tasks}
  	\end{center}
\end{figure*}

\subsubsection{Rotation Transformation}
In this experiment we consider MNIST images (\cite{lecun-mnisthandwrittendigit-2010}). \\ MNIST images consists of hand-written images of digits from 0-9 and the task is to train an image classifier which can classify images into appropriate digits. We modify the task slightly by rotating the validation and test images by 30\textdegree (this information is kept hidden) while the training images are not rotated. If we train a classifier on the un-rotated images, then it will perform poorly on the validation and test data set. Therefore, it becomes imperative to learn the hidden angle by which the validation and test images are rotated. We learn this hidden angle by minimizing the validation loss using BO. For the image classification task, we use LeNet \cite{le1989handwritten} as our base architecture and train it for 5 epochs. We compare the average validation loss over 50 runs. We run each of the BO algorithm for 30 iterations i.e. we train our CNN 30 times with different hyper-parameters. \textit{gPI} performs better than other algorithms (see Figure \ref{rotTrans}). Also, note that all algorithms have converged to $0$ loss.
\subsubsection{6D Regularization Problem}
    In this problem,  we minimize a validation loss function $f_{v}(x) = \sum_{i=1}^{6}(x_{i} - i + 0.5)^{2}$, where $x \in \mathbb{R}^{6}$. The training loss function used is $f_{T}(x) = \sum_{i=1}^{6}(x_{i}-10i)^{2} + ||\lambda^Tx||^{2}_{2}$, where $\lambda \in [0, 100]^{6}$ is a regularization coefficient which we learn by minimizing the validation loss using Bayesian optimization. The validation loss achieves the minimum value of $0$ at $\lambda=[19.00, 12.33, 11.00, 10.43, 10.11, 9.91]$. We compare the values of the validation loss function achieved by different BO algorithms. We run each BO algorithm for 200 iterations i.e. we query the objective function 200 times (excluding the $5$ initial queries for fitting the GP). We plot the average true validation loss over $200$ runs as a function of number of iterations. We observe from Figure \ref{6dreg} that \textit{gPI-MSC} outperforms most of the algorithms except for the FOBO algorithm.  Initially, FOBO's performance is poor, however, surprisingly it eventually outperforms other algorithms. Further research is warranted to understand why FOBO performs better than its counterparts in this setting. 

\begin{figure*}
 \begin{center}
      \subfloat[CartPole$-$v1]{
    \includegraphics[width = 5.5cm]{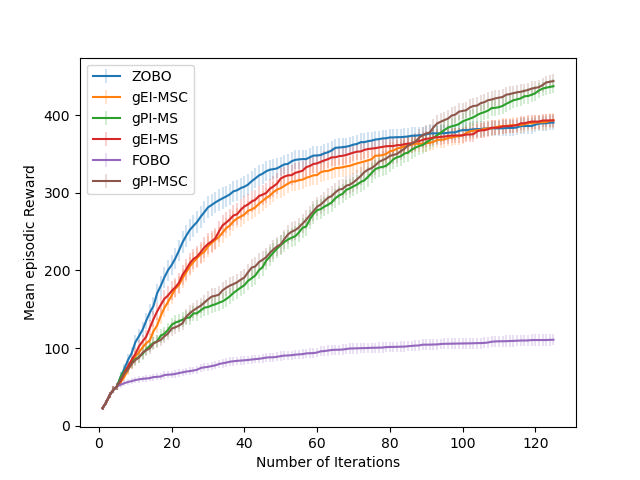}
    \label{cartpole}
    }
   \subfloat[GridWorld]{
  	\includegraphics[width = 5.5cm]{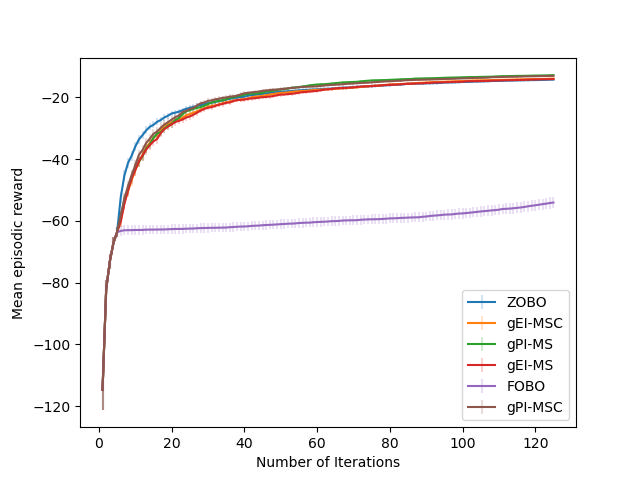}
  	\label{GridWorld}
  	}
  	% \subfloat[Gridworld]{
  	% \includegraphics[width = 5.5cm]{Regret_plot_gw.jpeg}
  	% \label{grid_world}
  	% }
  	\caption{RL Tasks}
  	\label{Tasks}
  	\end{center}
\end{figure*}

\subsection{RL tasks}
Readily available gradient information make RL an interesting domain to apply our \textit{gEI} and \textit{gPI} algorithms. We have primarily focused our attention towards integrating policy based methods with BO. Here, the y-axis correspond to mean episodic rewards. The estimates of gradient for policy based methods can be calculated (see supplementary material).
% \textcolor{red}{For obtaining the estimates of function, gradient and policy based methods, refer to \cite{prabuchandran2021novel} and our supplementary material.}
We consider Cartpole from the OpenAI gym and a custom grid world task to compare the performance of different BO algorithms. \cite{brockman2016openai}. 
\newline
\subsubsection{Cartpole} This task involves balancing a pole by suitably choosing between two actions (left and right) till the episode terminates. We have used a neural network to parameterize the policy. The architecture considered was a simple one layer neural network with 5 hidden units and with no bias. See Figure \ref{cartpole} for performance comparison of all the algorithms. While \textit{gPI-MSC} and \textit{gPI-MS} lag behind initially, towards the end they are able to outperform all the algorithms. The performance at the beginning can be attributed to the poor GP fitting and the choice of $\epsilon$ in our \textit{gPI} acquisition function. %This is left for future work.%which requires to be tuned further and is left for future work.

\subsubsection{Grid World} The objective of the agent is to find the most optimal path to exit the grid world using four actions (left, right, up, down). %Each state of grid world assigns a reward of -1 for each action. %and a solution can be easily found out by minimizing the time required to exit the grid world by finding the shortest path. 
The reward in a given state is sampled from an array of rewards [-5,-1,-2,1,2,5,10]. The sampling distribution is Dirichlet distribution and is fixed for each state. Thus, different states have different single-stage rewards. As a result, the shortest path (geometrically) in this grid world need not correspond to the optimal solution. For this problem, we use a one layer neural network with 5 hidden units and with no bias for parameterizing the policy. From Figure \ref{GridWorld}, we can see that our algorithms perform at par with ZOBO and outperform FOBO. The performance of our BO algorithms is not that significant over ZOBO and we suspect the choice of our kernel (RBF) do not suit here. Usually RL objective functions are not smooth and kernels as in \cite{wilson14} might prove to be more beneficial.

\section{Conclusions \& Future work}\label{conc}
In this work, we have proposed EI and PI based FOBO algorithms that have a superior performance compared to the state-of-the-art ZOBO and FOBO algorithms. The novelty in our algorithm lies in the design of the acquisition functions that can extract better information from the posterior GP models as compared to the existing FOBO algorithms. 

An important line of research would be to perform the regret analysis of the proposed algorithms on the lines of  \cite{srinivas2009gaussian}, \cite{shekhar2021significance}. Another interesting line of research would be to devise FOBO algorithms for time varying BO problems as in \cite{bogunovic2016time} and utilize FOBO algorithms in robust-optimization as in \cite{beland2017bayesian}. As our results suggests, no single acquisition function (based on EI or PI) can perform very well on all the black-box functions and in-fact we have a portfolio of acquisition functions, each of which can work well for suitable function choices. In such cases, it would be interesting to  perform a portfolio allocation procedure to suitably select the appropriate acquisition function each time as in \cite{hoffman2011portfolio}.

\newpage
\bibliographystyle{splncs04}
\bibliography{ecml}

\end{document}